\documentclass[final]{colt2020}
\usepackage{graphicx}
\usepackage{amssymb}
\usepackage{epstopdf}
\usepackage{natbib}
\usepackage{amssymb}
\usepackage{amsmath, mathabx}
\usepackage{mathrsfs}
\usepackage{hyperref}
\usepackage{comment}
\usepackage{xcolor}
\usepackage{enumitem}
\usepackage{algorithm}
\usepackage{algpseudocode}
\usepackage{stmaryrd}
\usepackage[utf8]{inputenc} 
\usepackage[T1]{fontenc}
\usepackage{fancyhdr}
\hypersetup{
    colorlinks=true,
    linkcolor=blue,
    filecolor=magenta,      
    urlcolor=cyan,
}
\newcommand{\A}{{\mathbf{A}}}

\newcommand{\bcA}{{\mathbfcal{A}}}

\newcommand{\bbP}{\mathbb{P}}

\newtheorem{Conjecture}{Conjecture}

\newtheorem{Challenge}{Challenge}

\DeclareMathAlphabet\mathbfcal{OMS}{cmsy}{b}{n}

\title{Open Problem: Average-Case Hardness of Hypergraphic Planted Clique Detection}%Is Hypergraphic Planted Clique Detection as hard as Planted Clique Detection?}
\coltauthor{%
 \Name{Yuetian Luo}$^\dagger$ \Email{yluo86@wisc.edu}
 \AND
 \Name{Anru R. Zhang}$^{\dagger *}$ \Email{anruzhang@stat.wisc.edu}\\
 \addr $^\dagger$Department of Statistics, University of Wisconsin-Madison\\ 
 $^*$Department of Biostatistics and Bioinformatics, Duke University
}

\date{\today}

\begin{document}
\maketitle

\setlength{\abovedisplayskip}{3pt}
\setlength{\belowdisplayskip}{3pt}

%%%%%%%%%%%%%%%%%%%%%%%%%%%%%%%%%%%%%%%%%%%%
\begin{abstract}
We note the significance of hypergraphic planted clique (HPC) detection in the investigation of computational hardness for a range of tensor problems. We ask if more evidence for the computational hardness of HPC detection can be developed. In particular, we conjecture if it is possible to establish the equivalence of the computational hardness between HPC and PC detection. 
\end{abstract}
%%%%%%%%%%%%%%%%%%%%%%%%%%%%%%%%%%%%%%%%%%%%

\section{Introduction}

The analysis of tensors has emerged as an active topic in recent years. Numerous datasets in the form of tensors emerge in various applications, such as collaborative filtering, neuroimaging analysis, hyperspectral imaging. In addition, tensor methods have been applied to many machine learning problems where the observations are not necessarily tensors, such as additive index models, high-order interaction pursuit, topic and latent variable models \citep{anandkumar2014tensor}. Compared with the inference for vectors or matrices, the tensor problems often possess distinct characteristics that bring significant computational challenges. As pointed out by the seminal work of \cite{hillar2013most}, 
extensions of many matrix concepts such as operator norm, singular values and eigenvalues are possible but computationally NP-hard. There often exist intrinsic gaps between the computational barriers and statistical limits in many tensor problems, as observed in tensor completion \citep{barak2016noisy}, tensor PCA \citep{richard2014statistical,hopkins2015tensor,wein2019kikuchi,zhang2018tensor,lesieur2017statistical,brennan2020reducibility}, high-order clustering \citep{luo2020tensor}, etc.  The analysis of computational barriers has attracted enormous attention because of its crucial role in the understanding of the computational feasibility of a wide range of tensor problems.

In the past decades, several lines of works have been devoted to the analysis of computational complexity in high-dimensional matrix problems. Such works include the average-case reduction \citep{berthet2013complexity}, analysis of restricted model of computation (e.g., statistical query \cite{kearns1998efficient}), analysis of algorithms such as sum-of-squares, belief propagation, approximate message passing, and analysis of optimization landscape. In particular, the average-case reduction requires randomized polynomial-time reduction from the conjectured hard problem in distribution to the target problem up to the conjectured computational barrier \citep{brennan2018reducibility}. Once the average-case reduction is established, all hardness results of the conjectured hard problem can be automatically inherited to the target problem. This provides a one-shot solution to the hardness of the target problem.

It would be ideal to use the same logic and establish the computational barriers for various tensor problems rigorously by average-case reduction from commonly raised conjectures, such as the planted clique (PC) detection or Boolean satisfiability problem (SAT), to the target tensor problem so that all of the hardness results of these well-studied conjectures could be naturally inherited. However, this route is often complicated by the multiway structure of the tensor. To overcome this, an alternative way of average-case reduction from hypergraphic planted clique (HPC) has been proposed in \cite{zhang2018tensor,luo2020tensor}. HPC has a more natural tensor structure that enables a more straightforward average-case reduction. A better understanding of the computational hardness relationship between PC detection and HPC detection can make a significant contribution to our knowledge of the average-case complexity in various tensor problems. 

\section{Planted Clique Detection and Hypergraphic Planted Clique Detection}
A $d$-hypergraph can be seen as an order-$d$ extension of regular graph. In a $d$-hypergraph $G = (V(G), E(G))$, each hyperedge $e \in E$ includes an unordered group of $d$ different vertices in $V$. Define $\mathcal{G}_d(N, 1/2)$ as the Erd\H{o}s-R{\'e}nyi $d$-hypergraph with $N$ vertices, where each hyperedge $(i_1, \ldots, i_d)$ is independently included in $E$ with probability $1/2$. Given a $d$-hypergraph $G$, define its adjacency tensor $\bcA = \bcA(G) \in (\{ 0,1 \}^N)^{\otimes d}$ as
\begin{equation*}
 	\bcA_{[i_1, \ldots, i_d]} = 
 	\begin{cases}
 	1, & \text{ if } (i_1, \ldots, i_d) \in E;\\
 	0, & \text{ otherwise}.	
 	\end{cases}
\end{equation*}
We define $\mathcal{G}_d(N, 1/2, \kappa)$ as the hypergraphic planted clique (HPC) model with the clique size $\kappa$. To generate $G \sim \mathcal{G}_d(N, 1/2, \kappa)$, we sample a random hypergraph from $\mathcal{G}_d(N, 1/2)$, pick $\kappa$ vertices uniformly at random from $[N]$, denote them as $K$, and connect all hyperedges $e$ if all vertices of $e$ are in $K$. The hypergraphic planted clique detection (HPC) can be formulated as the following hypothesis testing problem:
\begin{equation}\label{eq: HPC detection problem}
H_0: G \sim \mathcal{G}_d \left(N, 1/2\right)\quad \text{v.s.} \quad H_1: G \sim \mathcal{G}_d \left(N, 1/2, \kappa \right).
\end{equation}
We say a test $\phi$ solves the HPC detection if $\limsup_{N \to \infty} \bbP_{H_0} \left(\phi(\bcA) = 1\right) + \bbP_{H_1} \left(\phi(\bcA) = 0\right) = 0$, i.e., the sum of type-I, II errors converges to zero. 

When $d=2$, HPC detection reduces to the well-regard planted clique (PC) detection problem. The PC detection is statistically solvable if $\kappa \geq (2+\epsilon) \log_2 N$ for $\epsilon > 0$ by searching over all subgraphs of size $(2+\epsilon) \log_2 N$ \citep{bollobas1976cliques}. However, the best known polynomial-time algorithm requires $\kappa = \Omega(\sqrt{N})$ \citep{dekel2014finding,deshpande2015finding}. It has been widely conjectured that no polynomial-time algorithm can solve the PC detection when $\kappa = o(\sqrt{N})$. This conjecture has been strengthened in various types of computational models \citep{jerrum1992large,hopkins2018statistical,feldman2017statistical,deshpande2015improved,meka2015sum,barak2019nearly,gamarnik2019landscape}. As a result, the hardness conjecture of PC detection has become a key assumption in obtaining computational lower bounds for many problems, such as sparse PCA \citep{berthet2013complexity,wang2016statistical}, submatrix detection \citep{ma2015computational}, community detection \citep{hajek2015computational}, etc. 

Compared to PC detection, HPC detection ($d \geq 3$) has received far less attention in literature. \cite{bollobas1976cliques} proved that $K^d_N/\left(d!\log_2(N)\right)^{1/(d-1)} \overset{a.s.} \to 1$ if $K_N^d$ is the size of the largest fully connected subgraph in $G \sim \mathcal{G}_d(N, 1/2)$, which implies the HPC detection problem can be solved by exhaustive search in super-polynomial time when $\kappa \geq \left((d!+\epsilon) \log_2(N)\right)^{1/(d-1)}$ for any $\epsilon > 0$. \cite{zhang2018tensor} observed that applying SVD on the matricization/unfolding of $\bcA$ solves HPC detection efficiently if $\kappa = \Omega(\sqrt{N})$ but fails when $\kappa = N^{1/2 - \epsilon}$ for any $\epsilon>0$. Then they made a version of the following conjecture indicating that $\kappa = \Omega(\sqrt{N})$ is essential for solving HPC detection in polynomial time.
\begin{Conjecture}[HPC detection]\label{conj: hardness of tensor clique detection}
Suppose $d\geq 2$ is a fixed integer. If 
$$\limsup_{N \to \infty} \log \kappa / \log \sqrt{N} \leq 1 - \tau \quad \text{for any }\tau > 0,$$ 
for any sequence of polynomial-time tests $\{\phi \}_N: G \to \{0,1\}$, $\liminf_{N \to \infty} \bbP_{H_0} \left(\phi(G) = 1\right) + \bbP_{H_1} \left(\phi(G) = 0\right) > 1/2$. 
\end{Conjecture}
Recently, \cite{luo2020tensor} provides evidence for HPC detection conjecture by showing that \emph{Metropolis algorithms} \citep{jerrum1992large} and \emph{low-degree polynomial algorithms} \citep{hopkins2018statistical} can not solve the problem in polynomial-time when $\kappa = O(N^{\frac{1}{2} -\tau})$ for any $\tau > 0$.

\section{How hard is HPC detection?}

We raise the following challenges on the average-case computational hardness of HPC detection.

\vskip.1cm

\begin{Challenge}
	Can we establish the sharp limit of $\kappa$ to guarantee that HPC can be solved by SVD on the matricization/unfolding of $\bcA$? Is it possible to develop another polynomial-time algorithm for HPC that works below this limit?
\end{Challenge}

Such results for PC detection can be implied by \cite{knowles2013isotropic}  and \cite{deshpande2015finding}.

\begin{Challenge}
	 Can we provide more evidence for HPC detection conjecture under more classes of algorithms, such as statistical query (SQ), sum-of-squares (SOS), belief propagation (BP), approximate message passing (AMP), etc, in addition to the evidence in \cite{luo2020tensor}?
\end{Challenge}

\begin{Challenge}
	At the threshold level of $\kappa = O(N^{1/2-\tau})$ for $\tau > 0$, what is relationship between HPC detection and PC detection in their computational hardness? 
\end{Challenge}

First, HPC detection is no harder than PC detection. Given the adjacency tensor $\bcA$ of hypergraph $G$, by fixing $(d-2)$ indices and varying the remaining two, we get an adjacency matrix $\A$ from $\bcA$ which is a PC detection problem. So we can do a randomized polynomial-time reduction from HPC detection to PC detection without changing the planted clique size $\kappa$.

Now Challenge 3 becomes, how easier is HPC detection compared to PC detection? A one-shot approach to tackle this challenge is to do randomized polynomial-time reduction from PC to HPC. But this seems to be a very difficult task, upon which we have made several failed attempts. If reduction from PC to HPC could be proved, we would be able to arrive at the conclusion that PC and HPC detection are equivalently hard and all hardness results established in PC detection can be applied to HPC. The resulting connection between PC and HPC would enable solutions to a huge class of tensor computational hardness problems through existing matrix arguments. Such a connection can also provide solid evidence for the computational hardness assumptions of generalized planted clique problems proposed in \citep{brennan2020reducibility}, based on which we can establish various computational lower bounds more easily.

\bibliography{reference}

\end{document}